\documentclass[letterpaper, 10 pt, conference]{ieeeconf}  % Comment this line out if you need a4paper

\IEEEoverridecommandlockouts                              % This command is only needed if 
\usepackage{amssymb}
\usepackage{amsmath}
\usepackage{threeparttable}
\usepackage{booktabs}
\usepackage{algorithm}
\usepackage{algorithmicx}
\usepackage{algpseudocode}
\usepackage{xcolor}
\usepackage{graphicx}
\usepackage{subcaption}
\usepackage{multirow}
\usepackage{cite}    
\makeatletter
\let\NAT@parse\undefined
\makeatother
\usepackage{hyperref}
\hypersetup{
hypertex=true,
colorlinks=true,
linkcolor=blue,
anchorcolor=blue,
citecolor=blue
}

\title{\LARGE \bf
Jump-Start Reinforcement Learning with Self-Evolving Priors for Extreme Monopedal Locomotion
}

\author{Ziang Zheng$^{1}$, Guojian Zhan$^{1}$, Shiqi Liu$^{1}$,  Yao Lyu$^{1}$, Tao Zhang$^{3}$, Shengbo Eben Li$^{*1,2}$
\thanks{This study is supported by Tsinghua-Efort Joint Research Center for EAI Computation and Perception. $^{*}$All correspondence should be sent to Shengbo Eben Li. {\tt\small Email: lishbo@tsinghua.edu.cn}}% <-this % stops a space
\thanks{$^{1}$State Key Laboratory of Intelligent Green Vehicle and Mobility, School of Vehicle and Mobility, Tsinghua University, Beijing, China. $^{2}$College of AI, Tsinghua University, Beijing, China. $^{3}$SunrisingAI Ltd., Beijing, China.
}%
\thanks{}% <-this % stops a space
}

\begin{document}

\maketitle
\thispagestyle{empty}
\pagestyle{empty}

%%%%%%%%%%%%%%%%%%%%%%%%%%%%%%%%%%%%%%%%%%%%%%%%%%%%%%%%%%%%%%%%%%%%%%%%%%%%%%%%

\begin{abstract}
Reinforcement learning (RL) has shown great potential in enabling quadruped robots to perform agile locomotion. However, directly training policies to simultaneously handle dual extreme challenges, i.e., extreme underactuation and extreme terrains, as in monopedal hopping tasks, remains highly challenging due to unstable early-stage interactions and unreliable reward feedback. 
To address this, we propose JumpER (jump-start reinforcement learning via self-evolving priors), an RL training framework that structures policy learning into multiple stages of increasing complexity. By dynamically generating self-evolving priors through iterative bootstrapping of previously learned policies, JumpER progressively refines and enhances guidance, thereby stabilizing exploration and policy optimization without relying on external expert priors or handcrafted reward shaping. 
Specifically, when integrated with a structured three-stage curriculum that incrementally evolves action modality, observation space, and task objective, JumpER enables quadruped robots to achieve robust monopedal hopping on unpredictable terrains for the first time. 
Remarkably, the resulting policy effectively handles challenging scenarios that traditional methods struggle to conquer, including wide gaps up to 60 cm, irregularly spaced stairs, and stepping stones with distances varying from 15 cm to 35 cm. 
JumpER thus provides a principled and scalable approach for addressing locomotion tasks under the dual challenges of extreme underactuation and extreme terrains.
\end{abstract}

\section{Introduction}
\label{sec.intro}

% 1. 基本背景
%   1. 先讲四足机器人locomotion的背景（应用，结果）
%   2. 强化学习，大大地简化了实现的难度，能够真实地落地
% 2. 四足机器人两个问题，讲控制推向极限
%   1. 通过极限地形（复杂，稀疏接触）
%     1. 稀疏接触需要非常多的接触信息
%   2. 欠驱控制与极限模态
%     1. 单脚控制的这个模态是其中最困难的
% 3. 我们的工作内容
% 4. contribution

Quadruped robots have attracted increasing attention in recent years, emerging as highly versatile mobile platforms. 
Thanks to the flexibility of each leg to move independently, they are able to navigate challenging outdoor terrains~\cite{bellicoso2018advances, lindqvist2022multimodality}, perform agile maneuvers such as parkour jumps and backflips~\cite{cheng2024extreme}, and adapt to dynamic environments with notable robustness~\cite{zhang2024agile}. 
These achievements are primarily attributed to advancements in mechanical design and learning-based control algorithms. 
In particular, reinforcement learning (RL) has shown significant potential in simplifying controller design and conquering complex behaviors~\cite{li2023rlbook, cheng2025rambo, zhan2023enhance}.

\begin{figure}
    \centering
    \includegraphics[width=\linewidth]{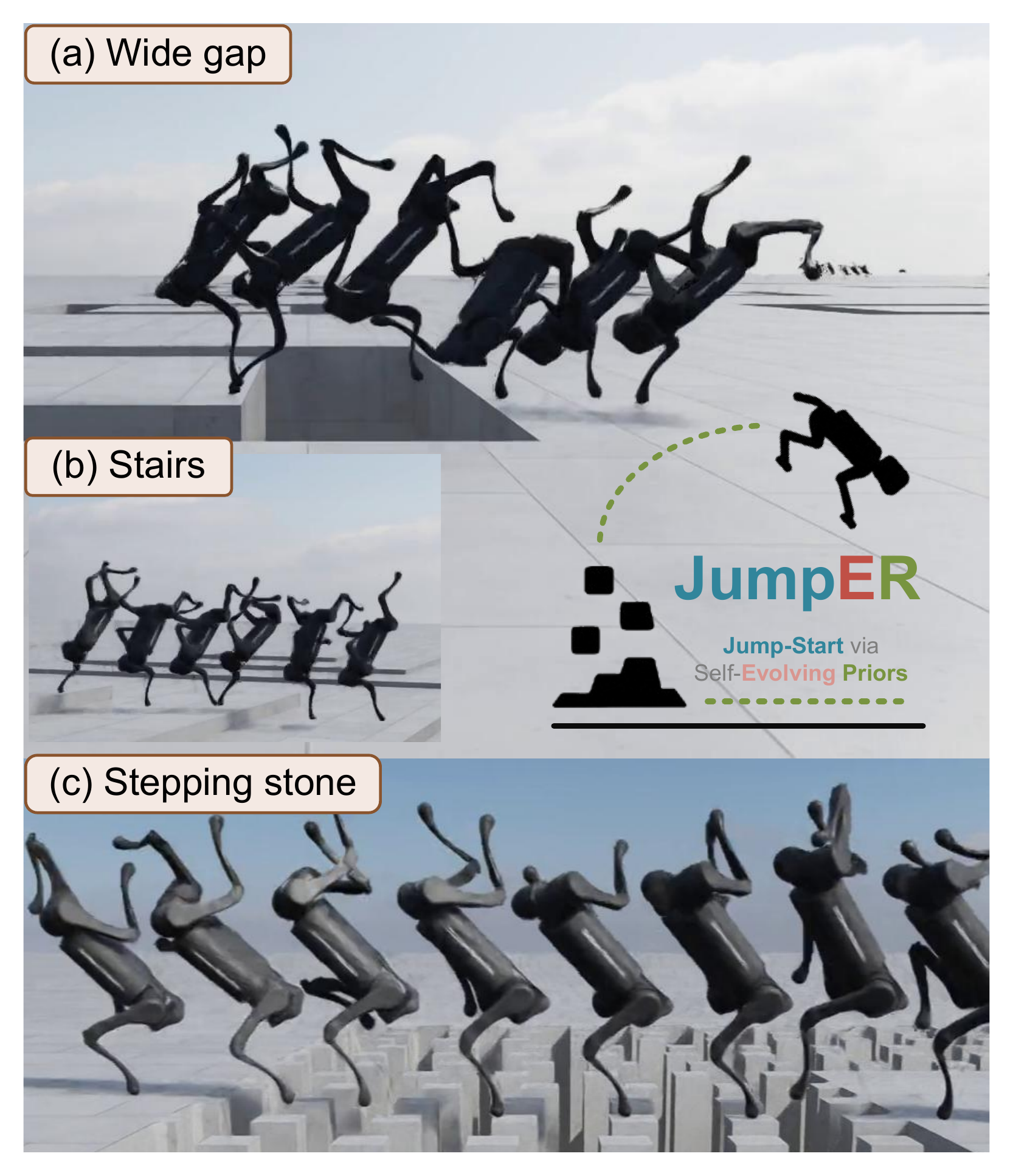}
    \caption{Monopedal hopping locomotion on extreme terrains tackled by JumpER.}
    \label{fig:intro_exp}
\end{figure}

Currently, quadrupedal locomotion on flat terrains has become relatively mature~\cite{zheng2025transferable}, enabling stable and efficient movements across uniform surfaces. However, significant challenges remain when robots encounter extreme terrains characterized by sparse, discontinuous, or unstable footholds, such as widely spaced gaps, stairs with varying heights~\cite{yu2024walking}, and stepping stones with uneven spacing (as shown in Fig. \ref{fig:intro_exp}). In these scenarios, robots are required to manage sparse and non-redundant contacts while carefully planning long-term foothold placements. 
Recent studies have started to address these challenges through various approaches, including terrain-aware locomotion over discrete footholds~\cite{miki2022learning}, hierarchical planning and control for parkour-like maneuvers~\cite{cheng2024extreme}, sim-to-real transfer for dynamic gap-jumping~\cite{lee2020learning}, and world model-based foothold prediction under irregular terrain~\cite{lai2024world}.

Although these methods improve performance on extreme terrains, they typically assume fully actuated systems with consistent ground contact and sufficient leg support. 
However, in practice, robots often face physical limitations, such as leg damage, mechanical constraints, or task-specific requirements that reduce the number of active legs. These conditions result in significant underactuation~\cite{tedrake2009underactuated}, causing highly nonlinear and unstable dynamics that conventional control methods struggle to manage~\cite{yang2024task}. 
Recent work has explored fault-tolerant locomotion strategies, particularly focusing on single or multiple leg motor failures~\cite{anne2021meta, hou2024multi, liu2024towards, xu2025acl}. 
Nevertheless, these approaches typically rely on maintaining consistent ground contact with at least two legs~\cite{li2024learning}, leaving the most extreme underactuated case, i.e., monopedal hopping with a single functional leg, largely unresolved.

Apparently, the combination of extreme terrain and extreme underactuation creates particularly challenging scenarios, referred to as dual extreme locomotion challenges. 
These scenarios are characterized by sparse and poorly shaped reward signals, leading to a fragile early-stage learning process. 
Standard RL methods often fail under these conditions, as initial policies struggle to initiate productive exploration, frequently collapse, and ultimately become trapped in local minima.
Recent efforts, such as the jump-start paradigm~\cite{uchendu2023jump, jiang2024rocket}, attempt to alleviate these challenges by introducing fixed prior policies at the beginning of each episode to guide policy learning, which avoids premature convergence to suboptimal behaviors and accelerates policy learning in tasks with delayed or sparse feedback. However, jump-start methods depend on existing nominal policies that can already roughly accomplish the desired task, which is usually unavailable in dual extreme locomotion scenarios.

In this paper, we propose JumpER ({jump}-start reinforcement learning via self-{e}volving p{r}iors), an RL training framework designed to tackle the dual challenges of extreme underactuation and extreme terrains, with a focus on achieving robust monopedal hopping locomotion on unpredictable terrains.\footnote{Code is available at: \href{https://anonymous.4open.science/r/JumpER-85B3}{\texttt{anonymous.4open.science/r/JumpER}}}
Our main contributions are as follows:

% These frozen priors provide structured guidance throughout training, shielding the agent from collapsing into local minima caused by sparse or misleading reward signals.
% Built on this framework, we design a progressive training scheme that enables a quadruped to transition from multi-leg locomotion to single-leg hopping across discontinuous and extreme terrain.
% We validate our framework across a range of tasks involving single-leg jumping on uneven, discontinuous, and sparse terrains. 
% As shown in Fig.~\ref{fig:intro_exp}, our method enables robust, agile locomotion that includes gap crossing, narrow foothold navigation, and dynamically stable monopedal hopping—capabilities rarely achieved in prior RL-based quadruped locomotion work.

\begin{itemize}
    \item JumpER introduces a multi-stage jump-start paradigm that leverages self-evolving priors. Unlike conventional fixed-policy guidance, JumpER dynamically generates priors through iterative bootstrapping of previously learned policies. Each training stage progressively refines and strengthens guidance, effectively avoiding premature convergence due to insufficient exploration and sparse rewards. This mechanism significantly stabilizes training and accelerates convergence towards the optimal policy.
    \item To the best of our knowledge, we are the first to achieve robust monopedal hopping of quadruped robots on extreme terrains. Our approach integrates a carefully structured three-stage curriculum explicitly designed for incremental refinement of policies in monopedal hopping tasks. By progressively scaling complexity through transforms in action modality, observation space, and task objective, the curriculum seamlessly aligns with JumpER's self-evolving priors, thereby facilitating continuous and effective policy improvement even under highly challenging conditions.
\end{itemize}

\begin{figure*}
    \centering
    \begin{subfigure}{\linewidth}
        \centering
        \includegraphics[width=\linewidth, trim=5 5 5 5, clip]{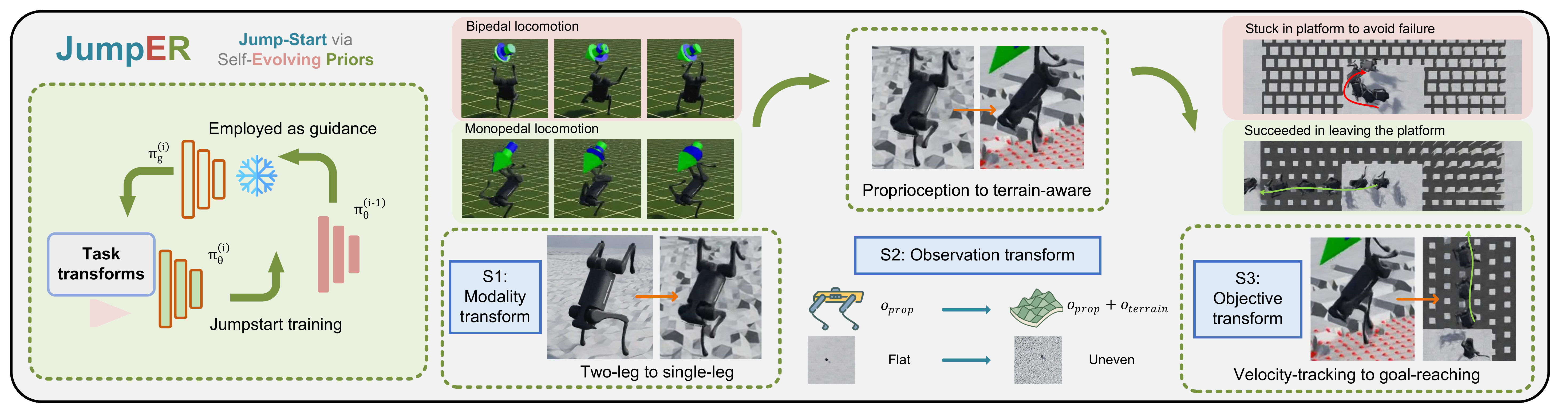}
        \caption{
            Three-stage training curriculum.
        }
        \label{subfig:three_stage}
    \end{subfigure}
    
    \begin{subfigure}{0.49\linewidth}
        \centering
        \includegraphics[width=\linewidth, trim=5 0 5 40, clip]{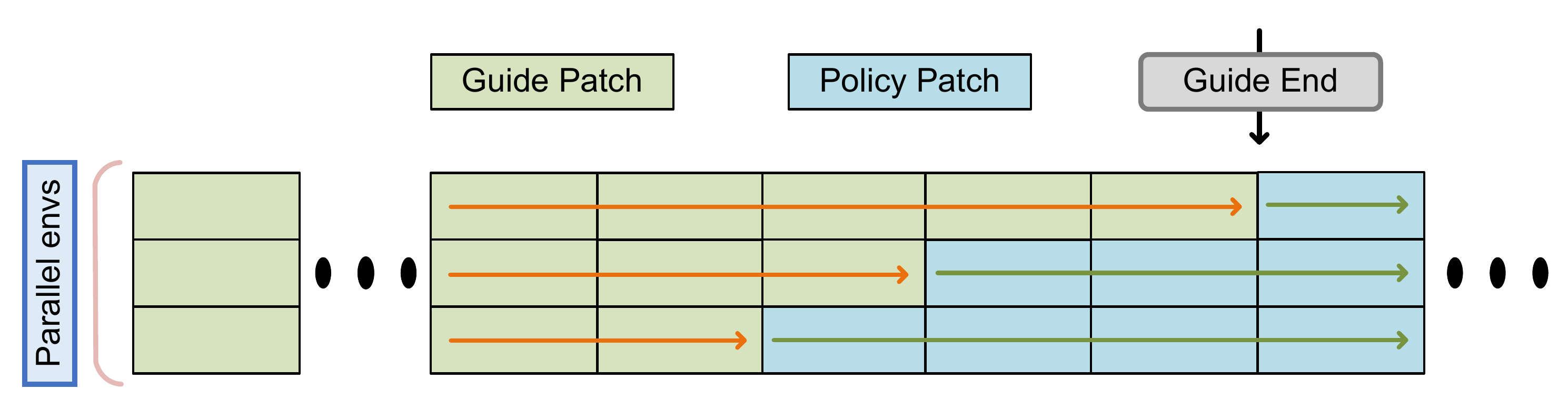}
        \caption{
            Patch-level jump-start sampling. 
        }
        \label{subfig:patch_jsrl}
    \end{subfigure}
    \begin{subfigure}{0.49\linewidth}
        \centering
        \includegraphics[width=\linewidth, trim=5 20 5 5, clip]{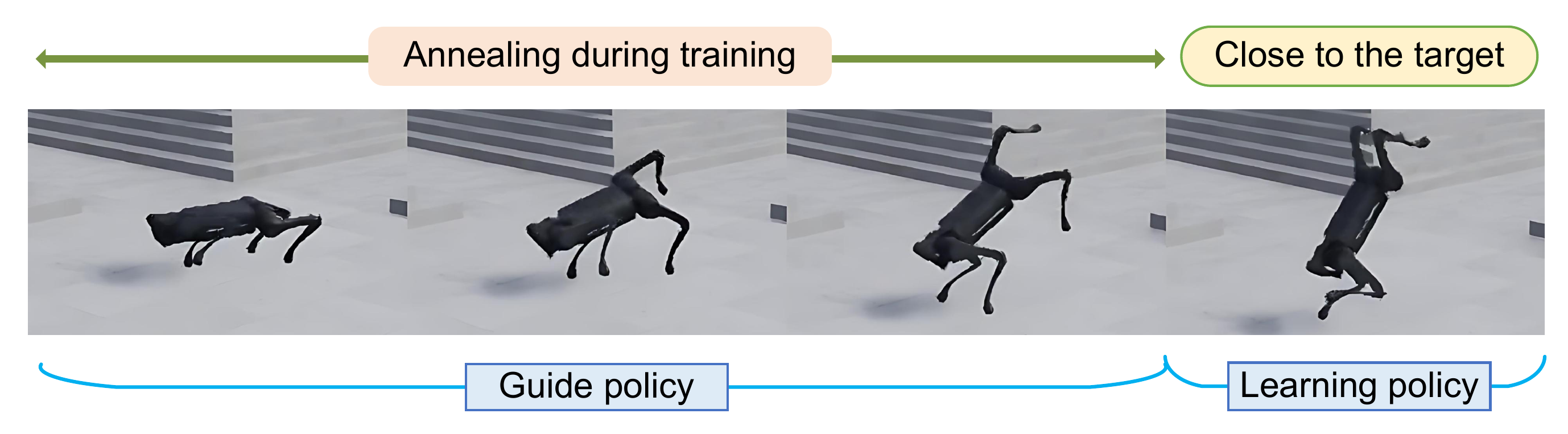}
        \caption{
            Annealing jump-start scheduling.
        }
        \label{subfig:patch_schedule}
    \end{subfigure}

    \caption{
        Overview of JumpER framework for achieving robust monopedal locomotion on extreme terrains. In the three-stage training curriculum, the task becomes increasingly difficult through deliberate transformations. 
            Stage 1 performs a \textit{modality transform}, transitioning from a partial bipedal prior to a monopedal policy. 
            Stage 2 introduces an \textit{observation transform}, augmenting proprioceptive inputs with terrain information to handle rough terrains. 
            Stage 3 performs an \textit{objective transform}, shifting the task objective from velocity tracking to goal reaching, guiding the policy to better leave suboptimal regions.
        % It features a bootstrap learning pipeline with a three-stage curriculum, where the influence of the guidance policy and the learning policy is balanced through the jump scheduling strategy.
    }
    \label{fig:framework}
\end{figure*}
\section{Preliminaries}
\subsection{Monopedal Hopping on Extreme Terrains}
As illustrated in Fig.~\ref{fig:intro_exp}, we conquer the task of monopedal hopping with a quadruped robot, an inherently unstable and highly dynamic setting. In this task, the robot must reach a target location using only one leg in contact with the ground at any given time, while the other legs remain lifted throughout. Such a constraint poses significant challenges for balance, control authority, and energy-efficient motion coordination.

To further increase difficulty, we consider extreme terrain conditions, including widely spaced gaps that require long-distance leaps, irregularly spaced stairs that disrupt rhythmic motion, and narrow stepping stones that demand precise foot placement. Together, these elements create a task landscape that pushes the limits of dynamic locomotion and real-time control.

\subsection{Reinforcement Learning}
RL considers a Markov decision process (MDP) denoted by $\mathcal{M} = (\mathcal{S}, \mathcal{A}, r, P, d_0, \gamma)$, where $\mathcal{S}$ and $\mathcal{A}$ represent the state and action spaces, $r: \mathcal{S} \times \mathcal{A} \rightarrow \mathbb{R}$ is the reward function, $P: \mathcal{S} \times \mathcal{A} \times \mathcal{S} \rightarrow \mathbb{R}_+$ is the transition probability function, $d_0$ is the initial state distribution, and $\gamma \in [0,1)$ is the discount factor~\cite{zhan2024transformation}. The objective of RL is to find a policy $\pi(a|s)$ that maximizes the expected discounted return, defined as
\begin{equation}
\label{eq.object_function}
J(\pi) = \mathbb{E}_{s_0 \sim d_0,\, a_t \sim \pi,\, s_{t+1} \sim P(\cdot \mid s_t, a_t)}\left[\sum_{t=0}^{T} \gamma^t r(s_t, a_t)\right],
\end{equation}
where $T$ denotes the length of the episode trajectory.
An episode typically ends when the agent reaches the target or when a predefined termination condition is triggered. In the monopedal hopping task, we define that any contact between the ground and any part of the robot, except for the single designated leg, will result in the termination of an episode.

Proximal policy optimization (PPO) ~\cite{schulman2017proximal} is a widely used on-policy RL algorithm in robot learning, valued for its ease of implementation, compatibility with parallel sampling, and training stability. 
% Due to these advantages, PPO has been widely adopted to tackle complex problems such as high-dimensional continuous control, professional-level game playing, and more recently, reinforcement learning from human feedback (RLHF) \cite{lambert2025reinforcement} for fine-tuning large language models.
The core of PPO lies in its clipped surrogate objective:
\begin{equation}
\label{eq:ppo_clip}
L(\theta) = \mathbb{E}_t \left[ \min \left( \rho_t(\theta) \hat{A}_t, \, \text{clip}(\rho_t(\theta), 1 - \epsilon, 1 + \epsilon) \hat{A}_t \right) \right],
\end{equation}
where $\rho_t(\theta) = \frac{\pi_\theta(a_t|s_t)}{\pi_{\theta_{\text{old}}}(a_t|s_t)}$ is the importance sampling ratio and $\hat{A}_t$ is the estimated advantage\cite{schulman2015high}. The hyperparameter $\epsilon$ controls the degree of clipping and consequently constrains the policy update to stay within a small trust region, thereby improving learning stability.

% Despite its success in various domains, standard PPO struggles to perform effectively in dual extreme locomotion scenarios, such as the single-leg hopping task considered in this paper, where instability and sparse rewards impede policy learning.

\begin{algorithm}[t]
\caption{JumpER}
\label{alg:jump}
\begin{algorithmic}[1]
\Require Initial policy $\pi^{(0)}$, number of stages $\tau$, switching step $h$, environment $\mathcal{E}$, replay buffer $\mathcal{B}$
\For{$i \gets 1$ to $\tau$}
    \State Freeze $\pi^{(i-1)}$ as guidance policy $\pi_g^{(i)}$
    \State Initialize new policy $\pi_\theta^{(i)}$
    \For{each iteration (until convergence)}
        \State Initialize environment $\mathcal{E}$ and get initial state $s_0$
        \For{$t = 0$ to $T-1$}
            \If{$t < h$}
                \State $a_t \sim \pi_g^{(i)}(a_t|s_t)$ \Comment{Follow guidance}
            \Else
                \State $a_t \sim \pi_\theta^{(i)}(a_t|s_t)$ \Comment{Switch to policy}
            \EndIf
            \State Execute $a_t$ in $\mathcal{E}$, observe $s_{t+1}$ and $ r_t$
            \State Store transition $(s_t, a_t, r_t, s_{t+1})$ in $\mathcal{B}$
        \EndFor
        \State Update $\pi_\theta^{(i)}$ using collected trajectories
    \EndFor
    \State $\pi^{(i)} \gets \pi_\theta^{(i)}$
\EndFor
\State \Return final policy $\pi^{(\tau)}$
\end{algorithmic}
\end{algorithm}

\section{Methodology}
\label{sec:mth}

\subsection{Jump-Start RL via Self-Evolving Priors}

Jump-start RL introduces an exploration-enhanced learning framework, where a guidance policy $\pi_g(a|s)$ (typically an expert, rule-based, or pretrained policy) assists the learning policy $\pi_\theta(a|s)$  in interacting with the environment during the initial steps of each episode~\cite{uchendu2023jump, jiang2024rocket}. 
Given an episode length $T$ and a switching step $h \leq T$, a mixed policy $\pi_{\text{mix}}$ first executes the guidance policy $\pi_g$ and then switches to the learning policy $\pi_\theta$ for subsequent steps:
\begin{equation}
\pi_{\text{mix}}(a_t|s_t) = 
\begin{cases}
\pi_g(a_t|s_t), & \text{if } t < h, \\
\pi_\theta(a_t|s_t), & \text{if } t \geq h.
\end{cases}
\label{eq.jumpstart}
\end{equation}

This initial guidance effectively stabilizes early iterations and facilitates informed exploration. However, in highly challenging settings, such as dual extreme locomotion characterized by severely underactuated dynamics and unpredictable terrain, traditional jump-start paradigm face remarkable difficulties. Fixed or predefined guidance policies often fail to cope effectively due to unstable early interactions and sparse or deceptive rewards, making them insufficient for sustaining long-term learning and adaptation.

To address these limitations, we propose JumpER, a jump-start variant featured with
multi-stage and self-evolving priors specifically tailored for dual extreme locomotion tasks. JumpER divides the learning intro multiple stages and introduces a novel self-guided bootstrapping mechanism that progressively refines guidance policies from the last stage rather than relying on a static external prior.

Starting from an initial policy $\pi^{(0)}$, JumpER iteratively proceeds through $\tau$ bootstrap stages. 
At each stage $i \in \{1, \dots, \tau\}$, the learning policy trained in the previous stage, denoted as $\pi^{(i-1)}$, is frozen and employed as the guidance policy $\pi_g^{(i)}$ for the learning policy $\pi_\theta^{(i)}$ at the current stage. This process dynamically forms a mixed policy:
\begin{equation}
\pi_{\text{mix}}^{(i)}(a_t|s_t) = 
\begin{cases}
\pi^{(i-1)}(a_t|s_t), & \text{if } t < h, \\
\pi_\theta^{(i)}(a_t|s_t), & \text{if } t \geq h.
\end{cases}
\end{equation}

This self-guided curriculum enables each training stage to incrementally build upon the competencies acquired in previous stages. Consequently, policies gradually master increasingly complex locomotion dynamics and challenging terrain conditions, thereby achieving enhanced generalization, stability, and resilience.

By introducing adaptive, internally generated priors rather than relying on fixed, external guidance policies, JumpER effectively scales to handle the severe challenges associated with extreme underactuation and unpredictable terrains. This multi-stage framework ensures policy learning efficiency and robustness throughout the entire training process.

In essence, JumpER establishes a progressive chain of dynamically refined policies, aiming to tackle the inherent challenges of dual extreme locomotion by employing self-generated and progressively evolving priors. The complete procedure is detailed in Algorithm~\ref{alg:jump}.

\subsection{Three-Stage Curriculum for Monopedal Hopping on Extreme Terrains}
To tackle the particularly challenging task of monopedal hopping for quadruped robots on extreme terrains, we implement the proposed JumpER framework through a structured three-stage bootstrap curriculum. 
This curriculum progressively transforms the action modality, observation space, and task objective, thereby enabling the policy to adapt reliably and robustly as task complexity increases.

\paragraph{Modality transform}
As discussed in Section~\ref{sec.intro}, directly training a monopedal hopping policy on extreme terrains is highly challenging due to severe underactuation, unstable footholds, and limited high-quality training samples. Notably, even achieving stable single-leg balance on flat ground remains non-trivial. 
To address this challenge, we introduce stage 1 of our structured curriculum, termed modality transformation, as illustrated in Fig.~\ref{subfig:three_stage}. We begin by training a stable bipedal standing policy \cite{cheng2024extreme}, which serves as the first-stage guidance policy $\pi_g^{(1)}$ within the JumpER framework. This policy enables the collection of abundant training data with stable postures on flat terrain. 
To explicitly encourage single-leg hopping, we introduce penalties for ground contacts involving more than one foot. 
In addition, by specifying a velocity during training, once stable single-leg balance is achieved, the robot naturally transitions into upward hopping to track the desired velocity.
Consequently, this training stage facilitates the successful training of a policy for robust monopedal locomotion on flat ground, denoted as $\pi_\theta^{(1)}$.

\paragraph{Observation transform}
The trained monopedal hopping policy from the last stage $\pi^{(1)}=\pi_\theta^{(1)}$, relies solely on proprioceptive observations $o_\text{prop}$. Although effective on flat terrains, these limited observational inputs significantly constrain performance on complex terrains, such as rugged and sloped ground. To ensure robust locomotion in challenging environments, additional terrain observations $o_\text{terrain}$ become essential. Thus, in stage 2, we adopt $\pi^{(1)}$ as the guidance policy $\pi_g^{(2)}$ to facilitate training of a terrain-aware monopedal hopping policy $\pi_\theta^{(2)}$.
% This training stage is conducted on six diverse and challenging terrain configurations. 
Although the guidance policy $\pi_g^{(2)}$ does not incorporate terrain information, it provides brief episodes of balance on rough surfaces before failing. Such episodes provide critical interactions that expose relationships between terrain features and robot dynamics. 
By leveraging this data, the resulting policy $\pi_\theta^{(2)}$, conditioned jointly on proprioceptive and terrain observations, learns to adjust its actions in response to terrain variations. Consequently, the robot achieves stable and robust monopedal hopping on complex and unpredictable terrains. 
Additional details regarding observation elements and terrain features are described in Section \ref{sec:exp}.

\paragraph{Objective transform}
The terrain-aware policy, denoted as $\pi^{(2)}=\pi_\theta^{(2)}$, is trained under a velocity-tracking objective, which is a common formulation to develop fundamental locomotion capabilities. 
However, this objective can lead to suboptimal behaviors in certain difficult scenarios, such as hopping across stepping stones separated by large gaps.
Specifically, the policy may converge to local optima that minimize failure (e.g., spinning in place to avoid falling down) rather than attempting gap-crossing~\cite{zhang2024agile}, as shown in Fig.~\ref{subfig:three_stage}.
To overcome this limitation, we introduce stage 3 of the curriculum, termed objective transformation, in which the task objective is reformulated from velocity tracking to explicit goal-reaching. 
Here, we utilize the second-stage policy $\pi^{(2)}$ as the guidance policy $\pi_g^{(3)}$, leveraging its potential to produce trajectories that occasionally succeed in crossing gaps. 
The resultant learning policy, denoted as $\pi_\theta^{(3)}$, is trained without explicit velocity-based rewards. Instead, it receives a reward signal based on proximity to a predefined terrain-based goal. 
As a result, the robot progressively acquires the capability to consistently hop across stepping stones and reliably reach the designated target location, completing the JumpER learning pipeline for dual extreme locomotion. 

% Framework
\subsection{Practical Techniques}

To further improve policy performance and training stability in dual extreme locomotion tasks, we introduce several practical techniques. Their underlying principles are broadly applicable across a wide range of challenging tasks, including but not limited to the monopedal hopping considered in this work.

\paragraph{Patch-level gradient computation}
To ensure compatibility with the high-throughput parallelized simulation environment provided by Isaac Sim \cite{makoviychuk2021isaac}, we adopt a patch-level gradient computation approach for efficient and scalable policy optimization.
In conventional RL, policy gradients are typically computed over complete episodes~\cite{li2023rlbook}. 
However, such episode-level gradients often suffer from high variance due to the stochasticity of exploration, resulting in unstable training. 
To mitigate this issue, we instead compute gradients based on a large number of short trajectory segments, i.e., ``patches'', collected from parallel environment rollouts, as illustrated in Fig.~\ref{subfig:patch_jsrl}.
Aggregating gradient estimates across these short patches helps reduce variance caused by individual trajectory differences, leading to more stable and reliable policy updates. Furthermore, this approach leverages the benefits of parallel exploration to enhance data diversity, which helps improve generalization and reduce the risk of premature convergence to suboptimal policies.

% \paragraph{Jump scheduling strategy}
% To adaptively balance the influence of the guidance policy and the learning policy throughout training, we introduce a jump scheduling strategy that gradually reduces reliance on the guidance policy as the learning policy improves. 
% This process is controlled by a dynamic scheduling coefficient $\alpha_t$, which decays linearly over training steps:
% \begin{equation}
% \alpha_t = \max\left(0, 1 - \frac{t}{T_{\mathrm{decay}}} \right),
% \label{eq:alpha_decay}
% \end{equation}
% where $T_{\mathrm{decay}}$ represents the decay horizon. As depicted in Fig.~\ref{subfig:patch_schedule}, rollouts generated by the guidance policy dominate the early training stages, helping the learning policy overcome the initial exploration bottleneck. As training progresses, control is gradually handed over to the learned policy $\pi_\theta$, thereby encouraging autonomous skill development and reducing dependency on either handcrafted or pretrained guidance.

\paragraph{Patch-based jump scheduling}
\label{para:patch_schedule}
To gradually transfer control from the guidance policy to the learning policy, we adopt a {patch-based jump scheduling strategy}, where the mixture of guide and learned policies is applied at the granularity of patches rather than at individual timesteps.

We will have $N$ patches for each episode with an initial guide patch count $n_0 \leq N$ assigned to the guidance policy, we progressively reduce the number of guide-controlled patches as training advances. 
At training step $t$, the number of patches using the guide policy is defined as:
\begin{equation}
n_t = \max\left(0, n_0 - \left\lfloor \frac{t}{m} \right\rfloor\right),
\end{equation}
where $m$ denotes the number of environment steps between scheduling updates. 
As depicted in Fig.~\ref{subfig:patch_schedule}, rollouts generated by the guidance policy dominate the early training stages, helping the learning policy overcome the initial exploration bottleneck. As training progresses, control is gradually handed over to the learned policy $\pi_\theta$, thereby encouraging autonomous skill development and reducing dependency on either handcrafted or pretrained guidance.

\begin{table}[t]
\centering
\caption{Detailed reward terms and weights.}
\label{tab:reward_values}
\begin{threeparttable}
\begin{tabular}{lll}
\toprule
\textbf{Reward} & \textbf{Description / Equation} & \textbf{Weight} \\
\midrule
\multicolumn{3}{l}{\textbf{Task terms}} \\
\midrule
$R_{\text{track\_lin\_vel}}$        & $\exp\left(-\frac{\|\min(v, v^{\text{cmd}}) - v^{\text{cmd}}\|^2}{0.25}\right)$               & 1.5 \\
$R_{\text{track\_ang\_vel}}$        & $\exp\left(-\frac{(\omega_{\text{yaw}} - \omega^{\text{cmd}}_{\text{yaw}})^2}{0.25}\right)$   & 0.5 \\
$R_{\text{termination}}$ & Termination & -200.0 \\
$R_{\text{out\_bound}}$ & Reach terrain bound & 200.0 \\
\midrule
\multicolumn{3}{l}{\textbf{Dense terms}} \\
\midrule
$R_{\text{out\_platform}}$ & $\mathbf{1}\{ |x| > 3\}$ & 10.0 \\
$R_{\text{reach\_far}}$ & $exp(-||x||)$ & -0.5 \\

\midrule
\multicolumn{3}{l}{\textbf{Posture terms}} \\
\midrule
$R_{v_z}$                           & $v_z^2$                                                                                       & -0.5 \\
$R_{\omega_{xy}}$                   & $\|\omega_{xy}\|^2$                                                                           & -0.05 \\
$R_{\text{orientation}}$            & $\|g - g_{target}\|^2$                                                                               & -1.0 \\
\midrule
\multicolumn{3}{l}{\textbf{Smoothness and contact terms}} \\
\midrule
$R_{\text{joint\_torques}}$         & $\sum_{j} \tau_j^2$                                                                           & $-1.0 \times 10^{-5}$ \\
$R_{\text{action\_rate}}$           & $\sum_{j} (a_t - a_{t-1})^2$                                                                  & -0.01 \\
$R_{\text{action\_smoothness}}$     & $\sum_{j} (a_t - 2a_{t-1} + a_{t-2})^2$                                                       & -0.01 \\
$R_{\text{joint\_power}}$           & $\sum_{j} |\tau_j \cdot \dot{q}_j|$                                                           & $-2.0 \times 10^{-5}$ \\
$R_{\text{joint\_acc}}$             & $\sum_{j} \ddot{q}_j^2$                                                                       & $-2.5 \times 10^{-7}$ \\
$R_{\text{joint\_deviation}}$       & $\sum_{j} (q_j - q_j^{\text{default}})^2$                                                     & -0.01 \\
% \midrule
% \multicolumn{3}{l}{\textbf{Basic Terms}} \\
% \midrule
$R_{\text{collision}}$              & $\sum_{i} \mathbf{1}\{F_i > 0.1\}$                                                            & -10.0 \\
$R_{\text{stumble}}$                & $\mathbf{1}\{\exists i, |F^{xy}_i| > 4|F^z_i|\}$                                              & -1.0 \\
$R_{\text{feet\_edge}}$             & $\sum_{i} c_i \sum_{d} \omega_d E_d[p_i]$                                                     & -1.0 \\
\bottomrule
\end{tabular}
\vspace{-3pt}
\end{threeparttable}
\end{table}

\begin{table*}[t]
\centering
\setlength{\tabcolsep}{4.0pt}
\caption{Performance comparison on sparse-reward locomotion tasks}
\label{tab:baseline-comparison}
\begin{threeparttable}
\begin{tabular}{ll|cc|cc|cc|cc}
\toprule
& & \multicolumn{2}{c}{Basic Posture} & \multicolumn{2}{c}{Monoped Posture} & \multicolumn{2}{c}{Objective Terms} & \multicolumn{2}{c}{Curriculum} \\
Task & Method & 
$\mathcal{R}_{\text{rear}} \uparrow$ & $\mathcal{P}_{\text{Base}} \downarrow$ & 
$\mathcal{R}_{\text{rear}} \uparrow$ & $\mathcal{P}_{\text{Mono}} \downarrow$ & 
$\mathcal{T}_{\text{Vel}} \uparrow$ & $\mathcal{T}_{\text{Reach}} \uparrow$ & 
Level $\uparrow$ & $\mathcal{R}_{\text{Success}}$ \\
\midrule

\multicolumn{10}{l}{\textbf{Stage 1: Modality Transform (flat terrains)}} \\
\midrule
\multirow{3}{*}{\textit{T1  balancing}} 
& Vanilla PPO       & 0.63 $\pm$ 0.08 & 0.24 $\pm$ 0.07 & 0.42 $\pm$ 0.06 & 0.10 $\pm$ 0.07 & -               & -   & -   & 88\% $\pm$ 9 \\ 
& PPO (pre)  & 0.68 $\pm$ 0.05 & 0.23 $\pm$ 0.04 & 0.93 $\pm$ 0.03 & 0.01 $\pm$ 0.02 & -               & -   & -   & 86\% $\pm$ 5 \\ 
& JumpER              &\textbf{0.71 $\pm$ 0.04} & \textbf{0.19 $\pm$ 0.03} & \textbf{0.95 $\pm$ 0.02} & \textbf{0.00 $\pm$ 0.01} & -               & -   & -   & \textbf{98\% $\pm$ 1} \\
\midrule  
\multirow{3}{*}{\textit{T2 upward hopping}} 
& Vanilla PPO       & 0.52 $\pm$ 0.07 & 0.45 $\pm$ 0.06 & 0.41 $\pm$ 0.07 & 0.54 $\pm$ 0.06 & 0.30 $\pm$ 0.05 & -   & -   & 22\% $\pm$ 5 \\ 
& PPO (pre)  & \textbf{0.64 $\pm$ 0.04} & 0.24 $\pm$ 0.03 & 0.52 $\pm$ 0.08 & 0.67 $\pm$ 0.14 & \textbf{0.57 $\pm$ 0.05} & -   & -   & 54\% $\pm$ 5 \\ 
& JumpER              & 0.60 $\pm$ 0.04 & \textbf{0.22 $\pm$ 0.03} & \textbf{0.94 $\pm$ 0.02} & \textbf{0.01 $\pm$ 0.01} & 0.53 $\pm$ 0.04 & -   & -   & \textbf{93\% $\pm$ 4} \\
\midrule

\multicolumn{10}{l}{\textbf{Stage 2: Observation Transform (uneven terrains)}} \\
\midrule
\multirow{3}{*}{\textit{T3 rough ground}} 
& Vanilla PPO       & -               & -               & -               & -               & -               & -               & 0.0 & -            \\ 
& PPO (pre)  & 0.67 $\pm$ 0.05 & 0.44 $\pm$ 0.13 & 0.66 $\pm$ 0.03 & 0.41 $\pm$ 0.09 & 0.23 $\pm$ 0.09 & 0.34 $\pm$ 0.06 & 3.3 & 40\% $\pm$ 5 \\ 
& JumpER              & \textbf{0.70 $\pm$ 0.05} & \textbf{0.20 $\pm$ 0.04} & \textbf{0.93 $\pm$ 0.03} & \textbf{0.07 $\pm$ 0.01} & \textbf{0.47 $\pm$ 0.03} & \textbf{0.80 $\pm$ 0.06} & \textbf{8.0} & \textbf{89\% $\pm$ 6} \\
\midrule
\multirow{3}{*}{\textit{T4 slope and stairs}} 
& Vanilla PPO       & -               & -               & -               & -               & -               & -               & 0.0 & -            \\ 
& PPO (pre)  & 0.65 $\pm$ 0.05 & 0.27 $\pm$ 0.04 & 0.90 $\pm$ 0.03 & 0.02 $\pm$ 0.01 & 0.51 $\pm$ 0.05 & 0.50 $\pm$ 0.06 & 4.5 & 55\% $\pm$ 3 \\ 
& JumpER              & \textbf{0.68 $\pm$ 0.04} & \textbf{0.22 $\pm$ 0.03} & \textbf{0.91 $\pm$ 0.02} & \textbf{0.00 $\pm$ 0.01} & \textbf{0.55 $\pm$ 0.04} & \textbf{0.78 $\pm$ 0.05} & \textbf{7.8} & \textbf{79\% $\pm$ 2} \\
\midrule

\multicolumn{10}{l}{\textbf{Stage 3: Objective Transform (extreme terrains)}} \\
\midrule
\multirow{4}{*}{\textit{T5 wide gap}} 
& PPO (dense)       & -               & -               & -               & -               & -             & -               & 0.0 & -            \\ 
& PPO (pre)  & 0.67 $\pm$ 0.05 & 0.24 $\pm$ 0.04 & 0.87 $\pm$ 0.13 & 0.13 $\pm$ 0.11 & -             & 0.11 $\pm$ 0.06 & 0.7 & 16\% $\pm$ 5 \\
& PPO (dense + pre) & 0.62 $\pm$ 0.11 & 0.50 $\pm$ 0.11 & 0.80 $\pm$ 0.02 & 0.31 $\pm$ 0.06 & -             & 0.41 $\pm$ 0.06 & 2.4 & 34\% $\pm$ 8 \\  
& JumpER              & \textbf{0.70 $\pm$ 0.05} & \textbf{0.20 $\pm$ 0.04} & \textbf{0.93 $\pm$ 0.03} & \textbf{0.07 $\pm$ 0.17} & -             & \textbf{0.80 $\pm$ 0.06} & \textbf{8.0} & \textbf{87\% $\pm$ 2} \\
\midrule
\multirow{4}{*}{\textit{T6 stepping stone}} 
& PPO (dense)       & -               & -               & -               & -               & -             & -               & 0.0 & -            \\ 
& PPO (pre)  & \textbf{0.68 $\pm$ 0.23} & \textbf{0.15 $\pm$ 0.04} & 0.79 $\pm$ 0.03 & 0.31 $\pm$ 0.15 & -             & 0.09 $\pm$ 0.06 & 0.4 & 13\% $\pm$ 2 \\ 
& PPO (dense + pre) & 0.65 $\pm$ 0.05 & 0.67 $\pm$ 0.19 & 0.44 $\pm$ 0.03 & 0.90 $\pm$ 0.27 & -             & 0.33 $\pm$ 0.03 & 3.7 & 44\% $\pm$ 5 \\  
& JumpER              & 0.58 $\pm$ 0.04 & 0.32 $\pm$ 0.03 & \textbf{0.91 $\pm$ 0.02} & \textbf{0.10 $\pm$ 0.05} & -             & \textbf{0.78 $\pm$ 0.05} & \textbf{7.8} & \textbf{79\% $\pm$ 4} \\
% \midrule

\bottomrule
\end{tabular}

\vspace{2pt}
\begin{tablenotes}\footnotesize
\item[*] Each result is mean ± std over 5 runs. All metrics are normalized to [0, 1] except Level [0–9] and $\mathcal{R}_{\text{Success}}$ [\%].
\item[*] The marker ``-'' means not applied.
\end{tablenotes}
\end{threeparttable}
\end{table*}

\begin{figure}
    \centering
    \includegraphics[width=0.95\linewidth, trim={1.0cm 0.0cm 29cm 0.0cm}, clip]{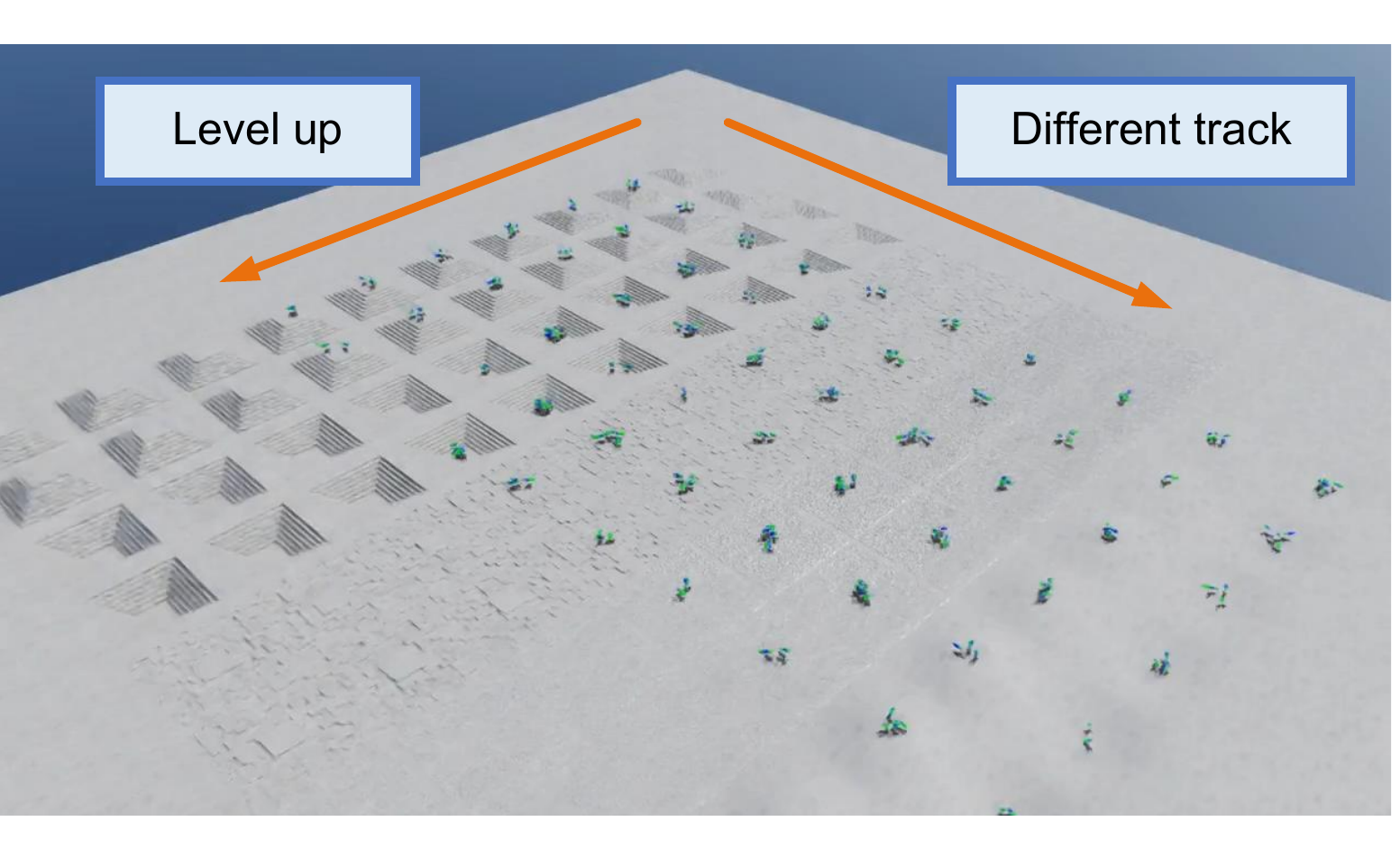}
    \caption{
    {Overview of the training playground, which includes terrains with diverse characteristics and difficulty levels.} 
    % Left: different tracks are arranged horizontally, while difficulty increases progressively in the vertical direction. 
    % Right: Illustrations for basic terrains.
    }
    \label{fig:playground}
\end{figure}

\section{Experiments}
\label{sec:exp}

In this section, we validate the effectiveness of our proposed JumpER framework by conquering monopedal hopping locomotion on extreme terrains. Precisely, we aim to answer the following key questions:

\begin{itemize}
    \item {Q1:} Does JumpER outperform conventional RL methods in performing monopedal hopping on both flat and uneven terrains? (Section~\ref{subsec:Q1_1} and \ref{subsec:Q1_2})
    \item {Q2:} Can JumpER further handle extreme terrain under a strict monopedal constraint? (Section~\ref{sec:exp-q2})
    % \item {Q3:} How does the guide policy in training help escape suboptimal policies? (Section~\ref{sec:q3})
\end{itemize}

% We evaluate across several highly dynamic locomotion scenarios, including both basic and extreme terrains.

\subsection{Experimental Setup}

We conduct our experiments using the IsaacLab framework \cite{mittal2023orbit}, with configurations adapted from IsaacGym \cite{makoviychuk2021isaac} to support large-scale, GPU-accelerated physics simulations.
All algorithms are implemented on top of RSL-RL~\cite{rudin2022learning, lyu2025conformal}.
The policy is trained under single-leg and double-leg task settings. 
In the single-leg scenario, the robot is restricted to using only one designated leg for locomotion. 
Any unintended contact, i.e., contact made by legs other than the designated one, results in immediate episode termination, encouraging precise and disciplined control.

\begin{figure*}[h]
    \centering
    \begin{subfigure}{0.49\linewidth}
        \centering
        \includegraphics[width=\linewidth, trim=5 5 5 5, clip]{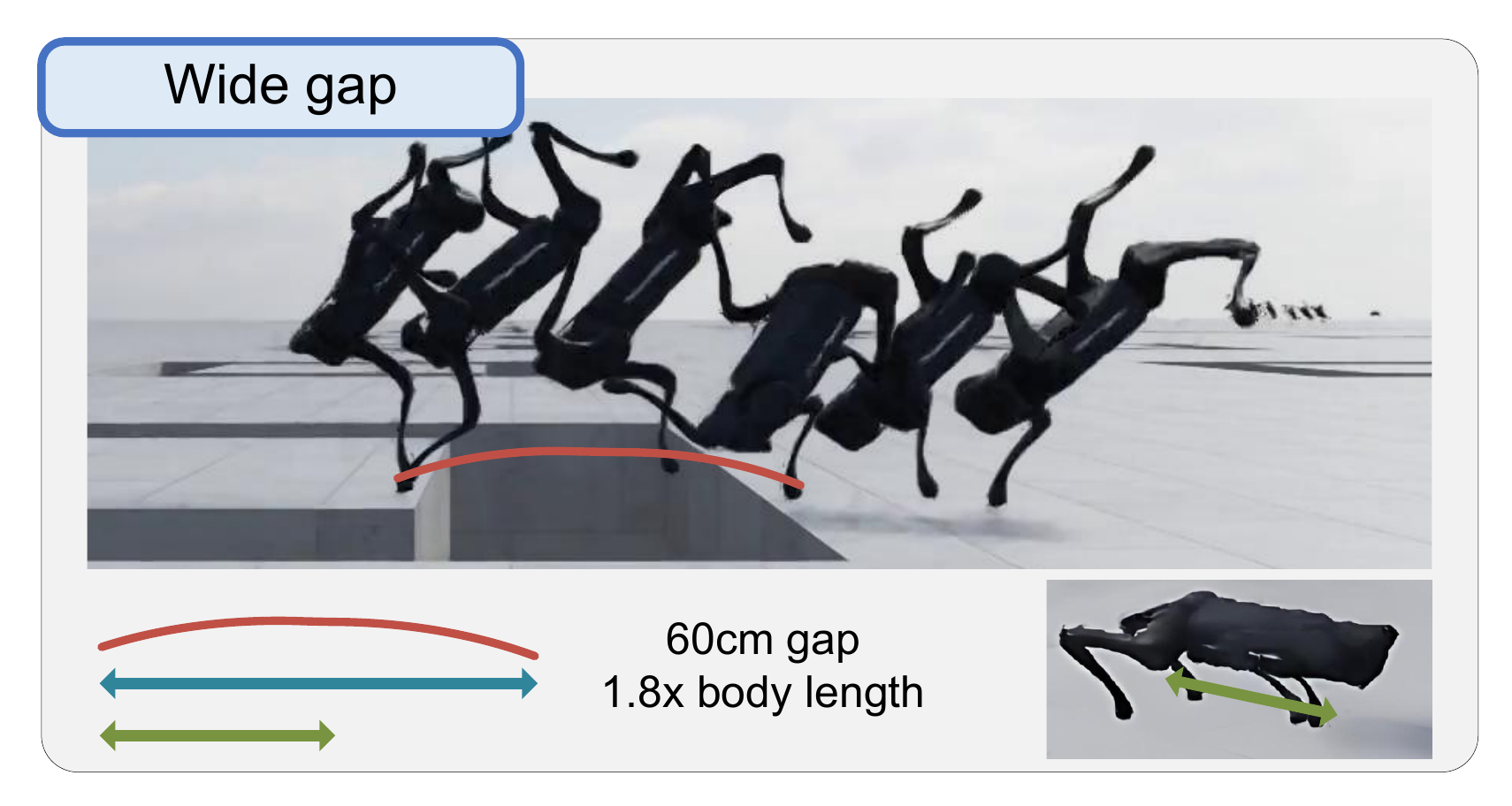}
        \caption{
        {Wide gap}. The robot performs dynamic leaps across varying gap distances up to 60 cm ($1.8\times$ body length).
        %     % A typical successful episode showing two distinct jump modes:
        %     % The robot must generate dynamic hops to traverse varying gap distances while maintaining balance with only a single leg. 
        %     % To the extreme, our policy (\textit{JumpER}) can perform long jumps over the gap for about 60 cm which is $1.2 \times$ of the body length and safely land on the platform.
        }
        \label{fig:exp_sketch_gap}
    \end{subfigure}
    \begin{subfigure}{0.49\linewidth}
        \centering
        \includegraphics[width=\linewidth, trim=5 5 5 5, clip]{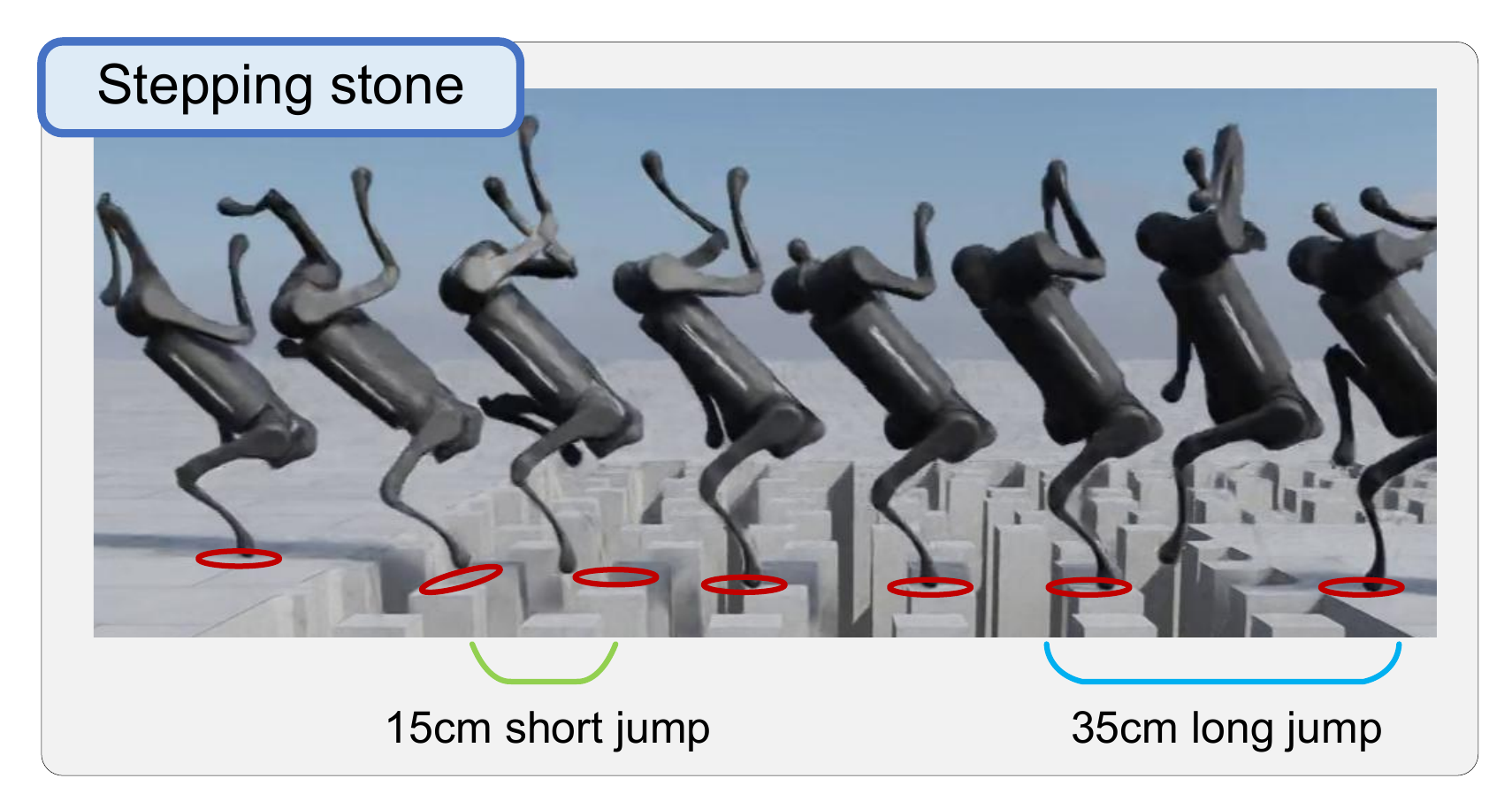}
        \caption{
        {Stepping stone}. The robot hops adaptively on stepping stones with jump distance ranging from 15 cm to 35 cm.
        %     % Illustration of the stepping stone terrain and the one-leg contact constraint. 
        %     % With a typical successful episode showing two distinct jump modes: 
        %     % (i) \textit{small jumps} precisely crossing single gap which is about 15 cm, and 
        %     % (ii) \textit{large jumps} spanning multiple gaps which is about 35 cm. 
        %     % For both jump will maintain a stability and precise contact.
        }
        \label{fig:exp_sketch_step}
    \end{subfigure}

    \caption{
        Visualization of monopedal hopping on extreme terrains.
    }
    \label{fig:diff_task_overview}
\end{figure*}

% \paragraph{Terrain benchmark}
We propose a comprehensive benchmark to evaluate locomotion capability in both basic and extreme terrains. 
The basic terrains are shown in Fig.~\ref{fig:playground}, while the extreme terrains comprise wide gaps and stepping stones (Fig.~\ref{fig:diff_task_overview}). 
All terrains are structured as curricula with gradually increasing difficulty. 
Each terrain is split into ten difficulty levels (detailed in~\cite{zheng2025transferable}); for the stepping stone task, we use a 2D curriculum with decreasing stone width and increasing gap spacing (Tab.~\ref{tab:stone_setting}). 
To progress through the curriculum, the agent must consistently succeed at its current difficulty level, where repeated failures will trigger a fallback to an easier stage. 

% \paragraph{Observation}
% We adopt the observation  structures from prior work~\cite{cheng2024extreme, zhang2024agile, zhan2024canonical, atanassov2024curriculum, zheng2025transferable}. 
% The policy inputs include the proprioceptive observations $o_{\rm prop}$, estimated base velocity $\hat{v}_t \in \mathbb{R}^3$, heightmap encodings around the feet and body $\hat{z}^f_t, \hat{z}^m_t \in \mathbb{R}^{16}$, and a latent vector $z_t$, while the policy outputs joint position actions $a_t \in \mathbb{R}^{12}$ tracked by PD controllers. Specifically, the proprioceptive observations $o_{\rm prop} \in \mathbb{R}^{45}$ includes:
% \begin{equation}
%     o_t = [\omega_t \quad g_t \quad cmd_t \quad \theta_t \quad \dot{\theta}_t \quad a_{t-1}]^\top
% \end{equation}
% where $\omega_t$, $g_t$, $cmd_t$, $\theta_t$, $\dot{\theta}_t$, and $a_{t-1}$ denote body angular velocity, gravity vector in the body frame, velocity command, joint angles, joint angular velocities, and the previous action, respectively. The $o_{\rm terrain}$ additionally receives terrain information:
% \begin{equation}
%     o_{\rm terrain} = [o_{\rm prop} \quad H^b_t \quad H^f_t]^\top
% \end{equation}
% where $H^b_t$, $H^f_t$ is the heightmaps around the body and feet. 

% \paragraph{Reward}
% \label{para:rew}
The reward function (TABLE~\ref{tab:reward_values}) promotes stable, efficient, and goal-oriented hopping~\cite{cheng2024extreme}, which comprises task, posture, and smoothness components. 
Task rewards alternate between velocity-tracking and goal-reaching based on curriculum progression.
We deliberately omit leg-height shaping from the reward, as the inherent vertical displacement during hopping introduces noise that destabilizes learning.  
For particularly challenging tasks in Stage 3, prior works~\cite{zhang2024agile} have introduced additional dense shaping terms that provide continuous positive rewards as the agent approaches the goal region. 
In our experiments, we adopt such dense rewards as a benchmark to facilitate comparison against our sparse-reward learning framework.

% \paragraph{Metric} 
We evaluate the performance from four aspects: (1)
For posture evaluation, we use two metrics: the base collision penalty $\mathcal{P}_{\text{Base}}$ for undesired body contacts, and the leg-on-air reward $\mathcal{R}_{\text{Air}}$, which measures the average suspension of rear legs and their distance from the target handstand height, both of them will be used in both handstand posture evaluation.
(2) For monopedal capability evaluation, we add to report the hand collision penalty $\mathcal{P}_{\text{Mono}}$ and the same leg-on-air reward $\mathcal{R}_{\text{Air}}$, now focused on the front leg’s ability to remain off the ground and near the desired height.
(3) For objective-specific performance, we use the return from velocity tracking $\mathcal{T}_{\text{Vel}}$ and goal reaching $\mathcal{T}_{\text{Reach}}$. 
(4) Finally, we report two curriculum metrics: the average achieved difficulty level ($\mathcal{L}_{\text{Level}}$), and the terrain success rate ($\mathcal{R}_{\text{Success}}$).

\begin{figure}[h]
\centering
\includegraphics[width=\linewidth, trim={0.0cm 0.0cm 0.0cm 0.8cm}, clip]{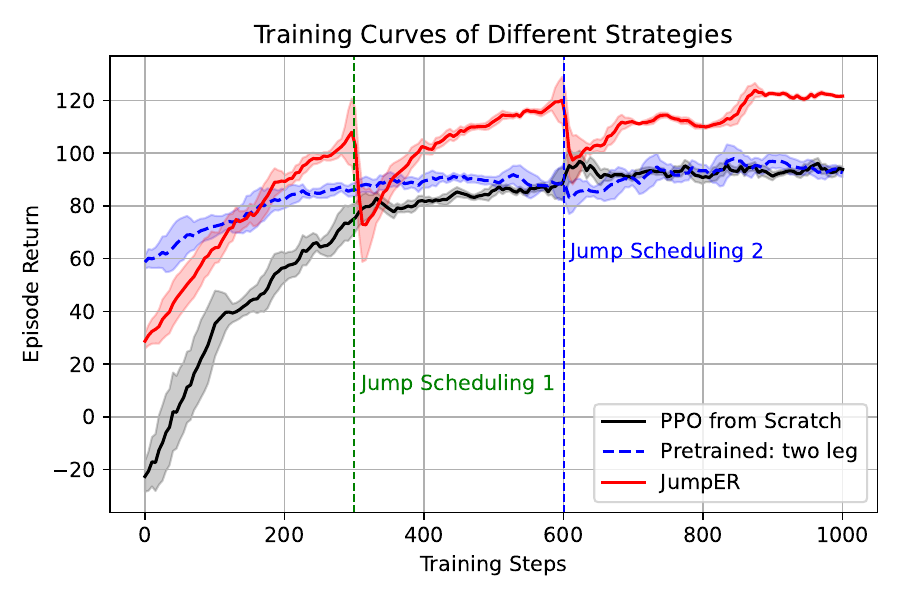}
\caption{{Training curves on \textit{Balancing} task.} Our JumpER (red) outperforms both scratch and naive pretraining baselines. The observed performance drops at 300 and 600 iterations are attributed to jump scheduling. This destabilization is promptly corrected by ongoing guidance and learning process.}
\label{fig:q3}
\end{figure}

\subsection{Stage 1: Monopedal Hopping on Flat Terrains}
\label{subsec:Q1_1}

We first evaluate whether JumpER can effectively solve sparse-reward locomotion tasks that are challenging for traditional RL methods. 
We compare against several strong baselines including: (1) {Vanilla PPO}~\cite{schulman2017proximal}: a standard on-policy RL algorithm trained from scratch. (2) {PPO (pretrained)}: PPO initialized from a pretrained policy. The learning policy is initialized from a two-leg training task.

% \begin{itemize}
%     \item {Vanilla PPO}~\cite{schulman2017proximal}: a standard on-policy RL algorithm trained from scratch.
%     \item {PPO (pretrained)}: PPO initialized from a pretrained policy. The learning policy is initialized from a two-leg training task.
% \end{itemize}

% \paragraph{Setup} 
We consider 4 representative terrains under the monopedal constraint, progressively increasing in difficulty and complexity. 
The first two are flat terrains for stage 1:
\begin{itemize}
    \item \textit{T1  balancing}: maintain a stationary upright pose without falling.
    \item \textit{T2 upward hopping}: repeatedly hop upward while maintaining a fixed horizontal position.
\end{itemize}

% \paragraph{Results}  
TABLE~\ref{tab:baseline-comparison} reports the metrics across all tasks. 
{JumpER} significantly outperforms baseline methods in basic flat terrain tasks (stage 1) and more difficult uneven terrain tasks. We visualize the training curves of JumpER alongside two naive baselines across on \textit{balancing} task in Fig.~\ref{fig:q3}.
We adopt $n_0 = 2$ guidance patches and a scheduling interval of $m = 300$ iterations for JumpER's patch-based transition strategy, where each patch spans 25 env steps and covers the critical stand-up phase.
The scratch baseline in black shows slow improvement and converges to a suboptimal solution, stuck due to poor exploration.
The pretrain baseline in blue improves initial performance but still converges suboptimally due to domain mismatch.
JumpER exhibits a clear pattern of stable learning: fast initial improvement, robustness through two bootstrapping transitions (around 300 and 600 iterations), and eventual convergence to a high-return policy. 
Performance temporarily degrades at predefined guide-policy phaseouts, where control transitions to the task policy. These handovers require the learner to reconstruct previously guided trajectory segments, resulting in short-term instability before regaining competence. 
Despite these transitions, the learning process quickly recovers and continues progressing toward the global optimum.

\subsection{Stage 2: Monopedal Hopping on Uneven Terrains}
\label{subsec:Q1_2}

% We first evaluate whether JumpER can effectively solve sparse-reward locomotion tasks that are challenging for traditional RL methods. 
We still use the two baselines: (1) Vanilla PPO and (2) PPO (pretrained), but now the pretrained policy is the best-performing policy in the last stage.
The two uneven terrains considered in this stage are
\begin{itemize}
    \item \textit{T3 rough ground}: move over stochastically generated uneven terrain at a given velocity.
    \item \textit{T4 slope and stairs}: move across inclined and stepped terrains with uneven height changes  at a given velocity.
\end{itemize}

We adopt the observation  structures from prior work~\cite{ atanassov2024curriculum, zheng2025transferable}. 
The policy inputs include the proprioceptive observations $o_{\rm prop}$, estimated base velocity $\hat{v}_t \in \mathbb{R}^3$, heightmap encodings around the feet and body $\hat{z}^f_t, \hat{z}^m_t \in \mathbb{R}^{16}$, and a latent vector $z_t$, while the policy outputs joint position actions $a_t \in \mathbb{R}^{12}$ tracked by PD controllers. Specifically, the proprioceptive observations $o_{\rm prop} \in \mathbb{R}^{45}$ includes:
$
    o_t = [\omega_t \quad g_t \quad cmd_t \quad \theta_t \quad \dot{\theta}_t \quad a_{t-1}]^\top
$
where $\omega_t$, $g_t$, $cmd_t$, $\theta_t$, $\dot{\theta}_t$, and $a_{t-1}$ denote body angular velocity, gravity vector in the body frame, velocity command, joint angles, joint angular velocities, and the previous action, respectively. The $o_{\rm terrain}$ additionally receives terrain information:
$
    o_{\rm terrain} = [o_{\rm prop} \quad H^b_t \quad H^f_t]^\top,
$
where $H^b_t$, $H^f_t$ are the heightmaps around the body and feet. 

% \paragraph{Results}  
TABLE~\ref{tab:baseline-comparison} reports that
{JumpER} also significantly outperforms baseline methods in more difficult uneven terrain tasks with terrain observation considered.

\begin{table}[h]
\centering
\caption{Uneven terrain configurations}
\label{tab:terrain_values}
\begin{threeparttable}
\begin{tabular}{llc}  % <-- 用 c 表示每列横向居中
\toprule
\textbf{Terrain type} & \textbf{Parameter} & \textbf{Value} \\
\midrule
\multirow{5}{*}{Slope and stairs}  & slope range     & (0.0, 0.2) \\
                               & slope platform width  & 2.0        \\
                               & stair height range & (0.05, 0.2) \\
                                & stair width        & 0.3         \\
                                & stair platform width    & 3.0         \\
\midrule
\multirow{2}{*}{Rough ground} & noise range & (0.02, 0.1) \\
                              & noise step  & 0.02        \\
\bottomrule
\end{tabular}
\vspace{3pt}
% \begin{tablenotes}\footnotesize
% \item[*] The range parameter is responsible for the terrain difficulty. Higher value means more difficult.
% \end{tablenotes}
\end{threeparttable}
\end{table}

\subsection{Stage 3: Monopedal Hopping on Extreme Terrains}
\label{sec:exp-q2}

To assess the robustness and adaptability of {JumpER}, we further consider two extreme terrains: 
\begin{itemize}
    \item \textit{T5 wide gap}: leap over a wide gap.
    \item \textit{T6 stepping stone}: pass through randomly spaced stepping stone.
\end{itemize}
These environments require precise foot placement, dynamic rebalancing, reliable contact transitions, etc. Such capabilities are especially difficult under sparse rewards and monopedal constraints.
Unlike previous stages focusing on velocity tracking, the task objective here is transformed to goal reaching, i.e., the agent must traverse extreme terrains to reach the goal located on the boundary.

\begin{table}[h]
\centering
\caption{Curriculum for extreme stepping stone.}
\label{tab:stone_setting}
\begin{threeparttable}
\begin{tabular}{cccc}
\toprule
\textbf{Difficulty level} & \textbf{Gap} & \textbf{Stone size} & \textbf{Stone gap} \\
\midrule
    0     & 10 cm    & 50 cm   & 5 cm \\
    \dots & \dots & \dots & \dots \\
    5     & 35 cm    & 25 cm   & 15 cm\\
    \dots & \dots & \dots & \dots \\
    9     & 60 cm    & 12.5 cm & 25 cm\\
% \midrule
\bottomrule
\end{tabular}
\vspace{3pt}
\end{threeparttable}
\end{table}

We further consider two stronger baselines with dense-shaped reward: (1) {PPO (dense)}: a reward-shaped baseline where the agent receives continuous positive reward proportional to its distance from the origin as defined in TABLE~\ref{tab:reward_values}. (2) {PPO (dense + pretrained)}: initialized from the best-performing policy from stage 2 (trained on basic terrain), then fine-tuned on extreme terrain with dense reachability rewards.

% \begin{itemize}
%     \item {PPO (dense)}: a reward-shaped baseline where the agent receives continuous positive reward proportional to its distance from the origin as defined in~\ref{para:rew}. This encourages forward progression.
%     \item {PPO (dense + pretrained)}: initialized from the best-performing policy from stage 2 (trained on basic terrain), then fine-tuned on extreme terrain with dense reachability rewards.
% \end{itemize}

% \subsubsection{Wide gap}

We provide visualizations in Figure~\ref{fig:diff_task_overview} to better show our effectiveness.
The policy trained with JumpER demonstrates emergent behaviors such as trunk tilting before jumping and mid-air posture adjustment, enabling it to jump over gaps with wide spacing up to 60 cm, as shown in Fig.~\ref{fig:exp_sketch_gap}. 
% Such behaviors are absent in PPO baselines, which fail to maintain balance or generate reasonable jump behavior.
For the stepping stone task, the robot must traverse irregularly spaced stepping platforms~\cite{zhang2024agile}. 
This setting requires both precise trajectory control and stability during brief, high-impact landings. 

% \begin{figure}[h]
% \centering
% \includegraphics[width=\linewidth, trim={0.0cm 0.0cm 0.0cm 0.8cm}, clip]{figures/exp/q3_ablation.pdf}
% \caption{{Training curves.} Our JumpER (red) outperforms both scratch and naive pretraining baselines. The observed performance drops at 300 and 600 iterations are attributed to jump scheduling. This destabilization is promptly corrected by ongoing guidance and learning process.}
% \label{fig:q3}
% \end{figure}

% \subsection{Training Curves}
% \label{sec:q3}

\section{Conclusion}

In this work, we present {JumpER}, a progressively bootstrapping learning framework tailored for dual extreme locomotion tasks. By leveraging a structured three-stage curriculum that gradually increases the difficulty in terms of modality, observation, and objective, JumpER successfully overcomes training instability and underactuation challenges by using dynamically improved priors. Through extensive experiments, we demonstrate that JumpER enables stable skill acquisition, and robust terrain adaptation in challenging monopedal hopping scenarios where conventional RL methods struggle, indicating practical utility in real-world applications. 

% Our results highlight the effectiveness of structured policy reuse and stage-wise curriculum in mastering dynamic, contact-rich locomotion tasks.

% \clearpage

\bibliographystyle{IEEEtran}
\bibliography{reference.bib}

\end{document}